\documentclass[letterpaper, 10 pt, journal, twoside]{ieeetran}

\usepackage[font=small,labelfont=bf]{caption}
\usepackage{multirow}
\usepackage{graphicx} 
\usepackage{caption}
\usepackage{subcaption}
\usepackage{url}
\usepackage{xcolor}
\usepackage{algorithm}
\usepackage[noend]{algpseudocode} 
\usepackage{amssymb,amsmath,comment,cite}
\usepackage{amsthm}
\usepackage{amsfonts}
\usepackage{cleveref}

\IEEEoverridecommandlockouts
\begin{document}
\newtheorem{theorem}{Theorem}
\newtheorem{lemma}{Lemma}
\newtheorem{definition}{Definition}
\def\mymethod{PSB}
\def\repolink{https://github.com/JingtianYan/PSB-RAL}

\markboth{IEEE Robotics and Automation Letters. Preprint Version. Jan, 2024}
{Yan \MakeLowercase{\textit{et al.}}: Multi-Agent Motion Planning with B\'ezier Curve Optimization under Kinodynamic Constraints}

\title{Multi-Agent Motion Planning with B\'ezier Curve Optimization under Kinodynamic Constraints}
\author{
Jingtian Yan$^{1}$, Jiaoyang Li$^{1}$


}
\maketitle

\begingroup
\let\clearpage\relax
    \begin{abstract}
Multi-Agent Motion Planning (MAMP) is a problem that seeks collision-free dynamically-feasible trajectories for multiple moving agents in a known environment while minimizing their travel time. 
MAMP is closely related to the well-studied Multi-Agent Path-Finding (MAPF) problem.
Recently, MAPF methods have achieved great success in finding collision-free paths for a substantial number of agents.
However, those methods often overlook the kinodynamic constraints of the agents, assuming instantaneous movement, which limits their practicality and realism.
In this paper, we present a three-level MAPF-based planner called PSB to address the challenges posed by MAMP. 
PSB fully considers the kinodynamic capability of the agents and produces solutions with smooth speed profiles. 
Empirically, we evaluate PSB within the domains of traffic intersection coordination for autonomous vehicles and obstacle-rich grid map navigation for mobile robots.
PSB shows up to 49.79\% improvements in solution cost compared to existing methods while achieving significant improvement in scalability.
\end{abstract}

    \section{Introduction}
\IEEEPARstart{M}{ulti-Agent} Motion Planning (MAMP) is a problem that focuses on finding collision-free dynamically-feasible trajectories for multiple agents in a known environment while minimizing their travel time.
MAMP has received significant attention in recent years, becoming a core challenge in various real-world applications, including traffic management\cite{li2023intersection},  warehouse automation\cite{honig2019warehouse}, and robotics\cite{sun2022multi}. 
MAMP is closely related to a well-studied problem called Multi-agent Path Finding (MAPF)\cite{Stern2019benchmark}, which plans collision-free paths in discrete timesteps for multiple agents on a given graph.
MAPF methods show the advantage of finding collision-free paths for hundreds of agents with optimal\cite{li2021eecbs} or sub-optimal\cite{li2021pairwise} guarantee.
However, the standard MAPF model does not consider the kinodynamic constraints of the agents. 
It assumes instantaneous movement and infinite acceleration capabilities, leading to discrete solutions that prescribe agents to move in synchronized discrete time steps and are thus not directly executable on their controllers.


\begin{figure}[!t]
\centering    \includegraphics[width=0.8\linewidth]{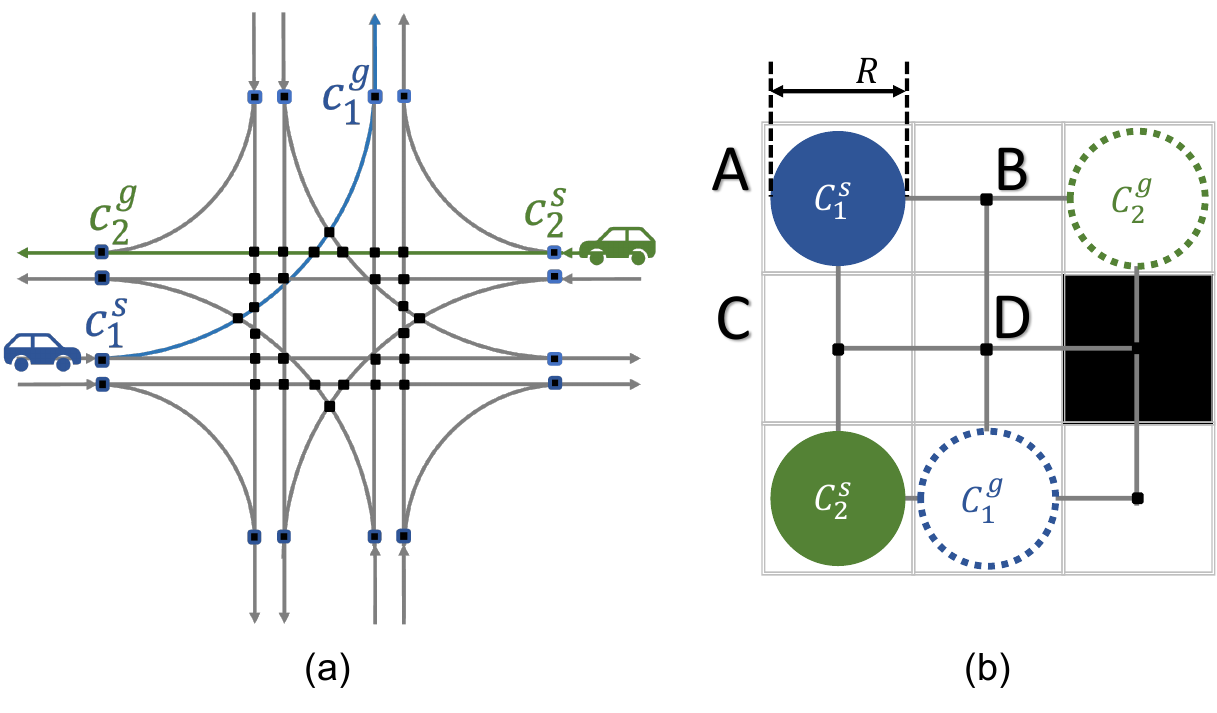}
    \caption{Illustration of the intersection model and the grid model we use. (a) Traffic intersection coordination model. (b) Grid map navigation model. Collision points are marked as black dots. The black cells in (b) are static obstacles.}
\label{fig:mapf_formulation}
\end{figure}

To tackle this challenge, some methods \cite{ali2023safe,solis2021representation,cohen2019optimal} discretize the action space and search using the motion primitives (i.e., elementary actions that an agent can perform).  
Those methods usually search in high-dimensional state space to account for the kinodynamics of agents such as speed and acceleration.
However, due to their discretized nature, their solutions do not fully capture the kinodynamic capacity of the agents, leading to limited choices of actions and, thus, poor performance in challenging scenarios.
To address this issue, a three-level method PBS-SIPP-LP (PSL)~\cite{li2023intersection}  is introduced for the traffic intersection coordination problem, a special case of MAMP. 
PSL integrates search-based and optimization methods. 
Level 1 and Level 2 employ an extended MAPF planner to identify potential paths, which invokes Level 3 to optimize vehicle travel speeds through a Linear Programming (LP) model.
However, PSL only considers speed constraints for agents, making a strong assumption that each agent moves through the intersection at a constant speed.




Taking inspiration from PSL, we introduce a three-level planner called PSB ({\textbf{\underline P}}BS-{\textbf{\underline {S}}}IPP-{\textbf{\underline B}}\'ezier) to address the MAMP problem.
Level 1 uses Priority-Based Search (PBS) \cite{ma2019searching} to resolve the collisions between agents.
Level 2 employs an extended Safe Interval Path Planning (SIPP) \cite{phillips2011sipp} to search candidate paths for each agent given constraints from Level 1.
Level 3 employs a B\'ezier-curve-based \cite{lorentz2012bernstein} planner to seek optimal speed profiles for agents based on the constraints from Level 2 and their kinodynamic constraints.
To reduce the runtime, PSB utilizes a cache structure to reuse solutions from Level 3. 
It also incorporates a duplicate-detection mechanism and a time-window mechanism to reduce the runtime further when extending PSB to the grid map navigation scenario.

The following content outlines our work in this paper: 
(1) We propose PSB, a MAMP planner capable of finding feasible, high-quality trajectories for multiple agents by exploiting their complete kinodynamic capability. 
PSB can produce speed profiles with any degree of smoothness. 
To our knowledge, this is the first MAPF-based work in MAMP that incorporates the full kinodynamic capability for the extensive group of agents.
(2) We show that PSB can efficiently handle the intersection coordination problem for autonomous vehicles with kinodynamic constraints (shown in Fig.~\ref{fig:mapf_formulation}(a)). Compared to the previously best algorithm PSL \cite{li2023intersection}, PSB shows up to 49.79\% improvement in terms of solution cost.
(3) We extend PSB to a more general domain: mobile robots navigation in obstacle-rich grid maps (shown in Fig.~\ref{fig:mapf_formulation}(b)). PSB outperforms the previously best algorithm SIPP-IP \cite{ali2023safe}, achieving up to 27.12\% improvement in solution cost and increased scalability to handle up to 220 agents, as opposed to the limit of 80 agents in SIPP-IP, while maintaining a comparable runtime.
    \section{Related Work}
Multiple algorithms have been developed to tackle MAMP.
Some methods extend single-agent motion planners for MAMP. In \cite{vcap2013multi}, a multi-agent RRT* derived from single-agent RRT*~\cite{karaman2011sampling} is proposed, combining the state space of individual agents into a collective joint space for RRT* planning. 
However, planning within the joint state space of agents presents scalability challenges, as the dimension of the joint state space increases exponentially in the number of agents.
Another category of methods uses control or optimization techniques to handle MAMP in real-world robotics. 
For instance, in~\cite{hegde2016multi}, a vector field approach combined with model predictive control is used to steer agents in complex environments. 
In~\cite{sun2022multi}, a mixed-integer linear program is utilized to search for timed waypoints that fulfill signal temporal logic constraints, which encapsulate the kinodynamic constraints.
These methods are effective in scenarios with few agents. 
However, they face scalability issues as the number of agents increases.
For instance, the approach by \cite{sun2022multi} takes over 100 seconds to generate the solution for only a four-agent scenario.

Recently, MAPF methods have achieved significant progress in finding discrete collision-free paths for a large number of agents.
Leading methods, such as Conflict-Based Search (CBS)\cite{sharon2015conflict, li2021eecbs} and Prioritized-Based Search (PBS)\cite{ma2019searching}, use single-agent solvers to plan paths for individual agents and address collisions among agents by introducing constraints to single-agent solvers. 
Some methods take advantage of the MAPF methods to solve the MAMP problem. 

One category of such methods uses discrete paths from MAPF planners and smooths them to meet kinodynamic constraints. 
For instance, the method in \cite{honig2016multi} enforces agents to adhere to their designated discrete paths generated by the MAPF planner.
Then, it considers the speed constraints and employs an LP solver to generate an executable plan.
Similarly, in \cite{zhang2021temporal}, agent trajectories, represented as B\'ezier curves, are optimized by accounting for high-order kinodynamic constraints using the paths from MAPF planners.
These methods highly rely on the discrete paths from MAPF planners, and their open-loop nature can lead to failures if kinodynamically feasible trajectories cannot be derived from these paths.

Another category of methods extends MAPF methods to consider robot kinodynamics.
For instance, CBS-MP \cite{solis2021representation} integrates probabilistic roadmaps with CBS to handle collisions and employs motion-primitive-based search for single-agent path planning.
Cohen et al.~\cite{cohen2019optimal} adapt CBS to continuous time and develop a bounded-suboptimal extension of SIPP for pathfinding for individual agents, where SIPP~\cite{phillips2011sipp} is a variant of $A^*$ that can find optimal paths in dynamic environments. 
Ali and Yakovlev \cite{ali2023safe} extend SIPP to account for the kinodynamics of agents by incorporating a wait interval projection mechanism, addressing the impractical assumption in SIPP of instantaneous stopping.
However, since these methods are all graph-search-based, they discretize the action space for search.
As a result, these methods consider a limited number of actions and thus fail to capture the full range of possible actions that agents could exhibit.

Recently, a three-level method called PSL~\cite{li2023intersection} is introduced, combining MAPF with optimization methods.
It uses PBS and SIPP to search for candidate paths and integrates an LP solver to optimize agent speeds along given paths.
However, PSL makes a strong assumption that each agent moves through
the intersection at a constant speed, which still fails to capture the full range of possible actions that agents could exhibit.
    \section{Problem Formulation} \label{sec:formulation}


We define our MAMP problem by a graph $G = (V, E)$ and a set of $M$ agents $\mathcal{A} = \{a_1, ..., a_M\}$.
Vertices in $V$, referred to as \textit{collision points}, represent potential collision locations for agents.
Agents can move from collision points $c_i \in V$ to $c_j \in V$ along edge $(c_i, c_j) \in E$ while adhering to kinodynamic constraints as shown in Eq.~\ref{eq:kinodymaic_constraints}, with the travel distance denoted as $d(c_i, c_j) \in \mathbb{R}^{+}$.
Each agent $a_i$ initiates its movement from a specified \textit{start (collision point)} $c_i^s \in V$ with initial kinodynamic constraints as shown in 
Eq.~\ref{eq:init} and moves toward a designated \textit{goal (collision point)} $c_i^g \in V$.
\begin{definition}
    (Path). The \textit{path} of agent $a_i$, if contains $m+1$ collision points, as $\phi_i = \{c_i^0=c_i^s, c_i^1,..., c_i^{m-1},c_i^m=c_i^g\}$ where $(c_i^{j-1}, c_i^j) \in E, j=1, \cdots, m$.
\end{definition}
\begin{definition}
    (Spatio-temporal profile). The \textit{spatio-temporal profile}, denoted as $\ell_i(t): \mathbb{R^+} \rightarrow \mathbb{R^+}$, quantifies the distance traversed by $a_i$ as a function of time $t$ along a given path.
\end{definition}
\begin{definition}
(Trajectory). The \textit{trajectory} of $a_i$ is the combination of a path and its associated spatio-temporal profile.
\end{definition}
The kinodynamic constraints limit the gradient of spatio-temporal profile with respect to time up to $K$-th order:
\begin{align}
\small
    \underline{U_i^k} \le {d^k \ell_i(t)}/{d t^k} \le \overline{U_i^k}, \forall k \in {1, ..., K} \label{eq:kinodymaic_constraints} \\
    {d^k \ell_i(t)}/{d t^k}|_{t=0} = \underline{U_i^k}, \forall k \in {1, ..., K} \label{eq:init}
\end{align}
where $\underline{U_i^k}$ and $\overline{U_i^k}$ are constant values that define the constraints on the $k$-th order gradient.
We use \textit{arrival time} $T_i$ to indicate the time needed for $a_i$ to arrive at $c_i^g$.
We denote $t_i(c)$ as the time when $a_i$ reaches the collision point $c$, and $\tau_{i}(c)$ represents the duration that it occupies $c$. 
Therefore, $a_i$ departs from $c$ at time $t_i(c) + \tau_{i}(c)$.
In the event that both $a_i$ and $a_j$ pass through the same collision point \textit{c}, a \textit{collision} occurs iff the time intervals $[t_i(c), t_i(c) + \tau_i(c))$ and $[t_j(c), t_j(c) + \tau_j(c))$ overlap.

\subsubsection{Intersection Model}
For the intersection coordination problem, we adopt the model from \cite{levin2017conflict}.
As shown in Fig.~\ref{fig:mapf_formulation} (a), the intersection has a set of entry lanes $\Gamma^{-}$ and a set of exit lanes $\Gamma^{+}$.
Each agent $a_i \in \mathcal{A}$, with a length of $L_i$, has a traveling request from the entry lane $c_i^s \in \Gamma^{-}$ to the exit lane $c_i^g \in \Gamma^{+}$ with the \textit{earliest start time} $e_{i} \in \mathbb{R}^{+}$ (i.e., the earliest time $a_i$ can reach at the entry lane).
To prevent agents from making sharp turns, they are constrained to follow the predefined path (gray lines in Fig.~\ref{fig:mapf_formulation} (a)).
We assume the agents can wait before the entry lane without any collision.
After entering the intersection, the minimum speed of agents is strictly positive.
If both $a_i$ and $a_j$ start from the same entry lane with $e_i < e_j$, then a \textit{overtake} happens if $\exists c \in V$ that $ t_i(c) \ge t_j(c)$.
Our task is to generate trajectories for all agents so that no collisions or overtakes happen while minimizing the sum of their arrival time.

\subsubsection{Grid Model}\label{sec:grid-def}
We adopt the grid model from classical MAPF problems~\cite{Stern2019benchmark} and represent $G$ as a four-neighbor grid map. Fig.~\ref{fig:mapf_formulation} (b) shows an example. There are three differences compared to the intersection model:
(D1) In contrast to the intersection model where there is only one path for each agent to move from its start to goal, the grid model involves an extra spatial domain search where we need to plan the path for each agent. 
(D2) Agents are allowed to stop at any location, which imposes a minimum speed constraint of $\underline{U_i^1}=0$.
(D3) All agents start simultaneously and remain at their respective goals after they finish.
It should be noted that this differs from the intersection model, where agents are only present in $G$ while in the intersection and not before entry or after departure.
Our primary objective is to plan collision-free trajectories for all agents while minimizing the sum of their arrival time.
    \section{MAMP on Intersection} \label{sec:method}
This section begins with a system overview of our proposed algorithm PSB, followed by a discussion of the specifics of our trajectory optimization formulation and method.
After that, we delve into the techniques used to tackle runtime challenges.
\begin{figure}[!t]
\centering
\includegraphics[width=0.82\linewidth]{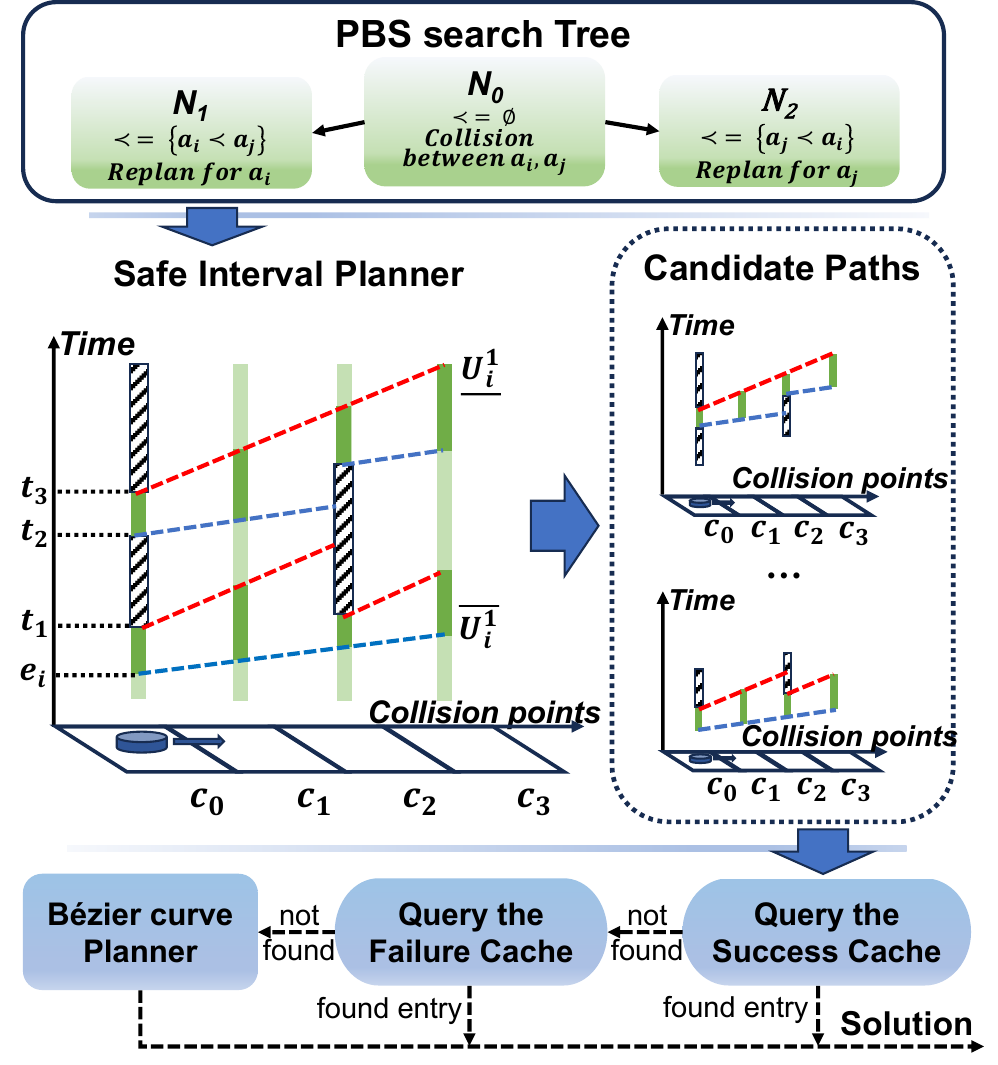}
    \caption{Overview of PSB.
    The shadowed strips denote time intervals occupied by other agents, the green segments denote safe intervals. The dark green segments represent the intervals in the open list.
    }
\label{fig:sipp_intersect}
\end{figure}

\subsection{System Overview}\label{sec:method_overview}
PSB consists of a three-level planner as illustrated in Fig.~\ref{fig:sipp_intersect}, with Level 1 and Level 2 extended from PSL~\cite{li2023intersection}.
Level 1 uses PBS \cite{ma2019searching} to resolve collisions among agents through priority ordering searching, where the trajectory of each agent is planned by Level 2.
Given the priority orderings from Level 1, Level 2 uses an extended SIPP~\cite{phillips2011sipp} to search for the optimal trajectory for an agent, where the spatio-temporal profile of the trajectory is optimized by Level 3.
Given the path (together with some temporal constraints) from Level 2, Level 3 uses BCP ({\textbf{\underline B}}\'ezier-{\textbf{\underline {C}}}urve-based {\textbf{\underline P}}lanner) to generate the optimal spatio-temporal profile.

\subsubsection{PBS-based planner}
We adopt PBS \cite{ma2019searching} to resolve collisions among agents.
If two agents have a priority ordering, the agent with lower priority must avoid collisions with the agent with higher priority. 
The goal is to find a set of priority orderings for agents so that they do not collide with each other.
To achieve this, we search a binary \textit{Priority Tree} (PT) in a depth-first manner. 
Each PT node contains a set of priority orderings and a set of trajectories consistent with the priority orderings.
We initialize the root PT node by adding priority orderings for the agents starting from the same entry lane based on their earliest start time to prevent \textit{overtake}.
Then, Level 2 is called to plan trajectories for agents at each lane from the highest priority to the lowest priority.
We expand PT nodes by searching for collisions between agents and generating child PT nodes with additional priority orderings to address the collisions.
For instance, as shown in Fig.~\ref{fig:sipp_intersect}, if a collision between agents $a_i$ and $a_j$ is detected in node $N_0$, we resolve it by expanding $N_0$ into two child nodes $N_1$ and $N_2$.
In $N_1$, $a_i$ is given higher priority than $a_j$, while in $N_2$, the priority is reversed.
Then, in each child node, we use Level 2 to replan trajectories for each agent based on the priority orderings in that node.
If we find a PT node with collision-free trajectories, we terminate the search and return the trajectories. 

\subsubsection{Safe Interval Planner}
Level 2 aims to find a trajectory for a given agent $a_i$ that minimizes its arrival time while avoiding collisions with higher-priority agents.
To achieve this, we run a modified SIPP on a safe interval graph, which associates each collision point with a set of safe intervals. 
A \emph{safe interval} $[lb, ub)$ is a time duration that an agent can stay at a collision point without colliding with higher-priority agents.
The modified SIPP plans a path (along with safe intervals) and calls Level 3 to specify the spatio-temporal profile considering kinodynamic constraints.
We provide an example of this process in Fig.~\ref{fig:sipp_intersect}, with a detailed description available in \cite{li2023intersection}.
While we use a 1-D example for demonstration, as later shown in the grid model, this search process is applicable in general graphs.
Since there are two safe intervals $[e_i, t_1)$ and $[t_2, t_3)$ at the start $c_0$, we generate two SIPP nodes and insert them into the open list.
In each iteration, we expand the node from the open list with the smallest $f$-value (= the lower bound of its safe interval plus the minimum arrival time required to reach the goal from its collision point).
In this example, we choose the node with safe interval $[e_i, t_1)$.
During node expansion, to speed up the search process while guaranteeing completeness, we use relaxed kinodynamic constraints. 
Specifically, we assume that $a_i$ occupies each collision point for only an instant of time while allowing adjustment of its speed within the range $[\underline{U_i^1}, \overline{U_i^1}]$ instantly. 
In our case, as $a_i$ moves from $c_i^{0}$ to $c_i^{1}$, a new interval $[e_{i} + t_{min}, t_{1} + t_{max})$ is generated at $c_i^{1}$, where $t_{min}=d({c_0}, {c_1})/\overline{U_i^1}$ and $t_{max}=d({c_0}, {c_1})/\underline{U_i^1}$ are the minimum and maximum time for this movement.
We iterate the safe intervals at $c_1$ that overlap with this new interval and insert them into the open list.
When a node reaches the goal, we backtrack to retrieve the full path and its associated safe intervals.
Then, we call Level 3 to determine the kinodynamically feasible trajectory.
If Level 3 finds a trajectory with a smaller arrival time than the best trajectory found so far, we update the current best trajectory.
We terminate the search if no node in the open list has a smaller $f$-value than the arrival time of the current best trajectory.
As shown in \cite{li2023intersection}, Level 2 guarantees to return the optimal trajectory when Level 3 is optimal and complete.

\subsubsection{B\'ezier-curve-based Planner}
Level 3 aims to generate a kinodynamically feasible spatio-temporal profile $\ell_i(t)$ for a given path within given safe intervals while minimizing the arrival time.
We use B\'ezier curve $B^{T_i}(t)$, where $T_i$ is the \textit{arrival time}, to represent $\ell_i(t)$ and reformulate this problem by finding the minimum $T_i$ and control points $P_i$ for $B^{T_i}(t)$ that satisfy the kinodynamic and safe-interval constraints.
As shown in Sec.~\ref{sec:method_bezier}, Level 3 is complete and optimal.

\subsection{Background on B\'ezier Curves}
A B\'ezier curve~\cite{lorentz2012bernstein} is a function parameterized by a set of control points. With a sufficiently large number of control points, it is able to approximate any continuous function $f(t)$ with $t \in [0, 1]$, making it an ideal choice for modeling $\ell(t)$~\cite{zhang2021temporal}.
We use $B^T(t)$ to denote the B\'ezier curve that scales the interval for $t$ from $[0, 1]$ to $[0, T]$:
\begin{equation}
\small
\label{eq:background_define_T_bernstein}
    B^T(t) = \sum_{r=0}^{n} p_{r}B_{r, n}^T(t), \; t\in[0,T],
\end{equation}
where $B_{r, n}^T(t) = \binom{n}{r}(\frac{t}{T})^r(\frac{T-t}{T})^{n-r}$ is called a Bernstein basis polynomial, and $P = \{p_0, ..., p_n\}$ are ${n + 1}$ control points.

As proven in \cite{lorentz2012bernstein}, the B\'ezier curve has three properties that can help find kinodynamically feasible trajectories efficiently:
(P1) $B^T(t)$ is bounded by the convex hull of its control points \textit{P} for $t\in [0, T]$. 
(P2) The curve always starts at the first control point ($B^T(0) = p_0$) and ends at the last control point ($B^T(T) = p_n$).
(P3) The derivative of the $n$-degree B\'ezier curve is another B\'ezier curve with degree $n-1$:
\begin{equation}
\small
     \frac{d^k B^T(t)}{d t^k} = \sum_{r=0}^{n-k} {p^{k}_{r}} B_{r, n-k}^T(t), \; t\in[0,T]
\end{equation}%
where ${p^{k}_{r}} = \frac{n!}{(n-k)!T^k}\sum_{j=0}^k (-1)^j \binom{k}{j} p_{n-j}$ is the $r$-th control point for the $k$-th gradient of $B^T(t)$. 

By properties (P1) and (P3), we can map the kinodynamic constraints discussed in Eq.~\ref{eq:kinodymaic_constraints} to constraints on control points: 
\begin{equation} \label{eq:bezier_gradient}
 \small
         \underline{U_{i}^{k}} \le {p^{k}_{r}} \le \overline{U_{i}^{k}}, \; \forall r\in\{0, 1..., n-k\}.
\end{equation}

\begin{figure}[!t] 
\centering
    \includegraphics[width=0.80\linewidth]{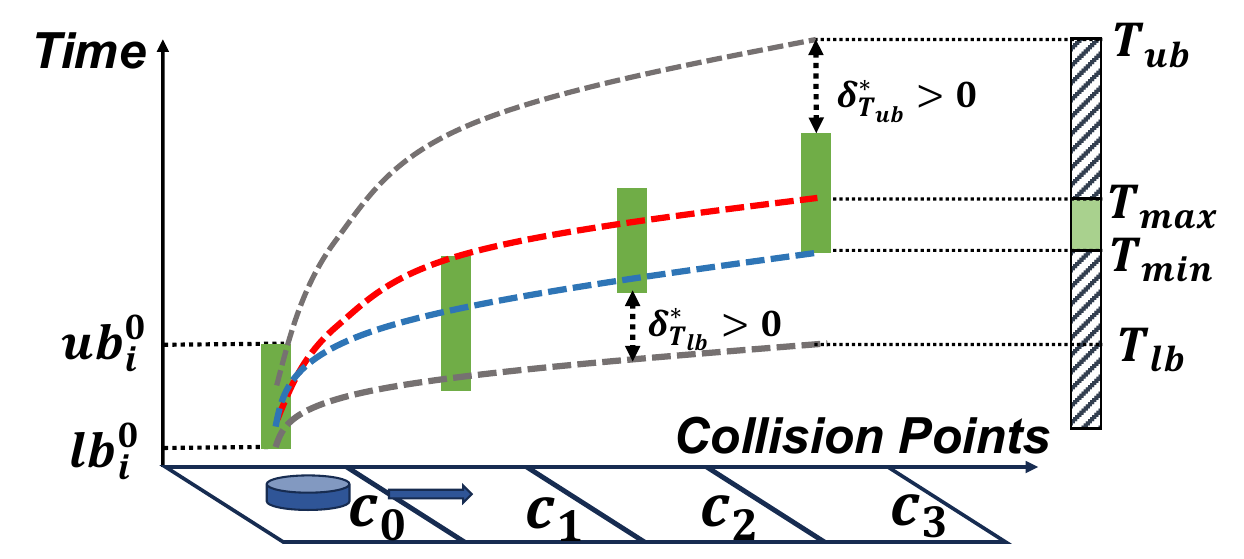}
    \caption{Illustration of the search for optimal arrival time. The agent moves along the path from $c_0$ to $c_3$ with associated safe intervals represented by green segments. The optimal spatio-temporal profiles found by Eq.~\ref{eq:bezier}-\ref{eqn:5} for the four different arrival time $T_x$ shown on the right are represented by four dashed curves. Since the two gray ones do not intersect with all green segments, they are infeasible, and $\delta^*_{T_{x}}$ indicates how far away they are from feasible profiles. As the figure shows, minimizing $\delta^*_{T_{x}}$ can bring these profiles closer to feasible ones.
    }
\label{fig:method_BCP}
\end{figure}

\subsection{B\'ezier-curve-based Planner (BCP)}\label{sec:method_bezier}
Given a path $\phi_i=\{c_i^0, \cdots, c_i^m\}$ from Level 2, along with its associated safe intervals $S_i = \{[lb_{i}^{0}, ub_{i}^{0}), \cdots, [lb_{i}^{m}, ub_{i}^{m})\}$, BCP aims to find a spatio-temporal profile $l_i(t)$, which we approximate by $B^{T_i}(t)$, for path $\phi_i$ within the safe intervals in $S_i$ that satisfies the kinodynamic constraints defined in Eq.~\ref{eq:kinodymaic_constraints}-\ref{eq:init} and minimizes the arrival time $T_i$.
Inspired by \cite{zhang2021temporal}, we introduce a method to determine $B^{T_i}(t)$ for a given arrival time $T_i$ and then a method to find the optimal arrival time $T_i^*$. In the rest of this section, we omit subscript $i$ for simplicity. 

\subsubsection{Find B\'ezier curve given arrival time}
Given arrival time $T$, our task is to determine the control points $P$ for $B^{T}(t)$  that meet all constraints, which we formulate as an LP problem:
\begin{align}
\small
\min_{P,\delta_T}\;  & \delta_{T} \label{eq:bezier}\\
s.t.\; & \delta_T \ge 0 \label{eqn:1}\\ 
     & p_0 = d^0,\quad p_n = d^m,\quad p_0^k = \underline{U}^k, \; \forall k \in \{1, ..., K\}\label{eqn:2}\\
     & B^{T}(lb^j) \le d^j + \delta_T, \; \forall j \in \{1, ..., m\}\label{eqn:3}\\
     & B^{T}(ub^j-L/\omega) > d^j + L - \delta_T, \forall j \in \{1, ..., m\} \label{eqn:4}\\
    & \underline{U^k}-\delta_T \le {p^{k}_{r}} \le \overline{U^k}+\delta_T, \label{eqn:5}\\
    & \hfill \quad \quad \quad \quad \forall k \in \{1, ..., K\}, \quad \forall r \in \{0, 1, ..., n-k \nonumber\}
\end{align}
where $d^j = \sum_{r=1}^{r=j}d(c^{r-1}, c^r)$ is the travel distance from start to collision point $c^{j}$.
Eq.~\ref{eq:bezier}-\ref{eqn:1} minimize $\delta_T$, a non-negative slack variable used to relax the safe-interval constraints and the kinodynamic constraints.  
$B^{T}(t)$ exists iff the optimal value for $\delta_T$ is zero.
Eq.~\ref{eqn:2} requires the agent to initiate from its entry lane with its initial kinodynamic constraints and terminate at its exit lane.
Given the challenge of finding a closed-form inverse function for $d^{j}=B^{T}(t)$,
Eq.~\ref{eqn:3}-\ref{eqn:4} ensure the agent visits each collision point within the safe interval by preventing arrivals before $lb^j$ (travel distance at $t=lb^j$ must be no larger than $d^j$) or departures after $ub^j$ (travel distance at $t=ub^j$ must be larger than $d^j+L$).
Here, $\omega$ is a constant derived from wave speed \cite{newell1993simplified} (we set it to 7 in our experiment).
Eq.~\ref{eqn:5} enforces the kinodynamic constraints from Eq.~\ref{eq:bezier_gradient}.

\begin{algorithm}[t]
\caption{B\'ezier-curve-based Planner (BCP)}
\label{algorithm:path-optimization}
\hspace*{\algorithmicindent} {\color{black} \textbf{Input: } Path $\phi$ and associated safe intervals $S$\\
\hspace*{\algorithmicindent} \textbf{Output: } The $\epsilon$-optimal $T^*$ and B\'ezier curve $B^{T^*}(t)$\\
\hspace*{\algorithmicindent} \textbf{Function} \textit{IsSolutionExist}$(T_{lb}, T_{ub})$\\
\hspace*{\algorithmicindent}\hspace*{\algorithmicindent} \textbf{return} $\nabla \delta_{T_{lb}}^*  \le 0\; and\; \nabla \delta_{T_{ub}}^* \ge 0$
}
\begin{algorithmic} [1]
{\color{black} \State $[T_{lb}, T_{ub}) \leftarrow [lb^m, ub^m)$
\If {$ub^m = \infty$} \quad $T_{ub} \leftarrow$ objective value of Eq.~\ref{eq:upper_bound} \label{line:upperbound}
\EndIf
\If{!\textit{IsSolutionExist}$(T_{lb}, T_{ub})$}\quad \Return{\textit{`no solution'}}
\EndIf
\While {$T_{ub} - T_{lb} > \epsilon$}
\State $T_{mid} \leftarrow (T_{lb}+T_{ub})/2$
\If{$\delta_{T_{mid}}^* = 0$}
\quad $T_{ub} \leftarrow T_{mid}$, $T^* \leftarrow T_{mid}$
\Else
\If{$\nabla \delta_{T_{mid}}^* > 0$}
\quad $T_{ub} \leftarrow T_{mid}$
\Else $\quad T_{lb} \leftarrow T_{mid}$
\EndIf
\EndIf
\EndWhile
\If{$T^* = null$} \quad \Return{\textit{`no solution'}}
\Else \quad \Return{$T^*$ and $B^{T^*}(t)$}
\EndIf}
\end{algorithmic}
\end{algorithm}

\subsubsection{Search for optimal arrival time}
To find the minimum arrival time $T^*$, BCP employs a binary search.
The LP model defined by Eq.~\ref{eq:bezier}-\ref{eqn:5} can be viewed as a function that maps an arrival time $T$ to the optimal slack value $\delta_T^*$. We denote the gradient of this function at $T$ by $\nabla \delta_T^*$, which, empirically, we approximate by $(\delta_{T+\Delta}^* - \delta_T^*)/\Delta$ with $\Delta$ being a sufficient small positive number. 
We borrow the following lemma from Theorem 2 of~\cite{zhang2021temporal}.
\begin{lemma}
\label{lemma:slack}
    $B^{T}(t)$ exists iff a single time range $[T_{min}, T_{max}]$ exists such that $\delta_T^* = 0$ for $T \in [T_{min}, T_{max}]$ and $\delta_T^* > 0$ for $T \notin [T_{min}, T_{max}]$, making $T_{min}$ the optimal travel time $T^*$. Moreover, $\nabla \delta_T^* < 0$ for $T \in [0, T_{min})$, $\nabla \delta_T^* = 0$ for $T \in [T_{min}, T_{max}]$, and $\nabla \delta_T^* > 0$ for $T \in (T_{max}, \infty)$. 
\end{lemma}
Thus, in an iteration of our binary search with $T^* \in [T_{lb}, T_{ub}]$, we can decide the search direction by evaluating the gradient of $\delta_T^*$ at the midpoint $(T_{lb}+T_{ub})/2$. 
The initial range $[T_{lb}, T_{ub}]$ for $T^*$ is set to the safe interval at the exit lane $[lb^{m}, ub^{m})$. However, due to the possibility of the agent waiting indefinitely at its entry lane, $ub^{m}$ can be infinite.
In that case, we compute $T_{ub}$ by the following LP model:
\begin{equation}
\begin{aligned}
\small
\label{eq:upper_bound}
\min_t  &\quad t + (d^m + L)/{\underline{U}}\\
s.t. &\quad lb_j \le t + d^j /\underline{U},
     \quad \forall j \in \{1, ..., m\}
\end{aligned}
\end{equation}
\begin{lemma}
\label{lemma:exist}
    If the latest arrival time at the exit lane is unbounded ($ub^{m}=\infty$), then there exists a valid $B^{T}(t)$ with $T$ being the objective value of Eq.~\ref{eq:upper_bound}.
\end{lemma}
\begin{proof}
{Since the minimum speed of agents is strictly positive, any safe interval with a finite upper bound restricts subsequent safe intervals along the path to also have finite upper bounds.
Therefore, $ub^{m}=\infty$ implies $ub^{j}=\infty$ for $j=0, \cdots, m$. 
As a result, the agent can wait at its entry lane for sufficient time and then maintain a continuous movement at $\underline{U}$ throughout the intersection. 
Eq.~\ref{eq:upper_bound} computes the minimum arrival time for such a solution.
}   
\end{proof}

\subsubsection{Pseudocode for BCP}
As shown in Algorithm~\ref{algorithm:path-optimization}, we initialize the range for $T^*$ by the last safe interval $[lb^m, ub^m)$ [Line 1].
In case of $ub^{m} = \infty$, the upper bound $T_{ub}$ is computed by Eq.~\ref{eq:upper_bound} [Line~\ref{line:upperbound}].
Then, we determine if $\exists T \in [T_{lb}, T_{ub})$ such that $B^T(t)$ exists. According to Lemma~\ref{lemma:slack}, $B^T(t)$ exists when $[T_{lb}, T_{ub})$ overlaps with $[T_{min}, T_{max}]$, which is true if $\nabla \delta_{T_{lb}}^* \le 0$ and $\nabla \delta_{T_{ub}}^* \ge 0$ (as shown in function \textit{IsSolutionExist}).
If $B^T(t)$ does not exist, BCP returns \textit{`no solution'} [Line 3].
Otherwise, we proceed with the binary search to find $T^*$ within range $[T_{lb}, T_{ub})$, and this search continues until the range is smaller than a predefined threshold $\epsilon$
(we use $\epsilon = 0.1$ in our experiments) [Line 4].
In each iteration, we average $T_{lb}$ and $T_{ub}$ to get $T_{mid}$ and solve Eq.~\ref{eq:bezier}-\ref{eqn:5} at $T_{mid}$ [Line 5].
If $\delta_{T_{mid}}^*=0$, it indicates that $T_{mid}\in[T_{min}, T_{max}]$. 
Consequently, we use $T_{mid}$ as the new upper bound for $T^*$ [Line 6].
On the other hand, if $\delta_{T_{mid}}^* \ne 0$, we calculate its gradient. 
If $\nabla \delta_{T_{mid}}^* > 0$, it implies that $T_{mid} > T_{max}$, and we thus use $T_{mid}$ as the new $T_{ub}$ [Line 8].
Conversely, it can be concluded that $T_{mid} < T_{min}$, and we thus use $T_{mid}$ as the new $T_{lb}$ [Line 9]. 

\begin{theorem}[Completeness and optimality of BCP]\label{lemma:BCP}
    Given a small enough $\epsilon$ and a large enough number of control points, BCP finds the spatio-temporal profile $B^{T^*}(t)$ with minimum arrival time $T^*$ if one exists and returns failure otherwise.
\end{theorem}
\begin{proof}
    With sufficient control points, the B\'ezier curve $B^T(t)$ is able to approximate any continuous function in $[0, T]$.
    Thus, Eq.~\ref{eq:bezier}-\ref{eqn:5} always find $P^*$ given $T^*$, if a feasible spatio-temporal profile exists.
    At the same time, with Lemma~\ref{lemma:exist}, the upper bound for $T^*$ is always finite. So after a finite number of iterations, 
    as shown in Theorem 3 of \cite{zhang2021temporal}, the binary search in BCP can find the $\epsilon$-optimal $T^*$.
\end{proof}

\begin{theorem}
[Completeness and suboptimality of PSL] PSB guarantees to find a (sub-optimal) solution in finite time.
\label{theo:comlexity}
\end{theorem}

\begin{proof}
    PSL~\cite{li2023intersection} guarantees to find a (sub-optimal) solution in finite time. The primary distinction between our PSB and PSL lies in Level 3. As Theorem~\ref{lemma:BCP} demonstrates that Level 3 of PSB is both complete and optimal, we can reuse the proof for PSL without modifications to show that our PSB also guarantees to find a (sub-optimal) solution in finite time. 
\end{proof}


\subsection{Caching BCP Results} \label{sec:method_interval}
As BCP encounters recurrent LP solving, calling BCP frequently can lead to a significant increase in runtime.
To address this issue, we implement two types of caches, namely \textit{SuccessCache} and \textit{FailureCache}, for each agent to utilize the results of previous BCP calls to expedite the search process.

While, in the intersection model, there exists only one path that moves each agent from its start to goal, we consider the general case where each agent can have multiple paths so that the caching mechanisms can be later applied to the grid model with no changes.
We say two paths for the same agent are \textit{pseudo-identical} iff they contain the same number of collision points and the travel distances between corresponding collision points are the same. Since the LP model in Eq.~\ref{eq:bezier}-\ref{eqn:5} uses the distances between collision points instead of the specific location of each collision point, we can relax the requirements for reusing results from both caches from the path being identical to pseudo-identical.

\textit{SuccessCache} maps paths with associated safe intervals to optimal spatio-temporal profiles.
During the search process, if a path is pseudo-identical to the path of an entry from \textit{SuccessCache}, and their associated safe intervals are identical, then we can reuse the spatio-temporal profile from that entry. 

In contrast, \textit{FailureCache} stores paths with associated safe intervals for which no spatio-temporal profiles exist. Notably, if no spatio-temporal profiles exist for a path with associated safe intervals, reducing the ranges of some safe intervals will still result in the non-existence of spatio-temporal profiles. 
Thus, during the search process, if a path is pseudo-identical to the path of an entry from \textit{FailureCache}, and each of its safe intervals is either identical to or a subset of the corresponding safe interval from the entry, we can infer that no spatio-temporal profiles exist. 



As shown in Fig.~\ref{fig:sipp_intersect}, during the search process of Level 2, after obtaining a path $\phi_i$ with associated safe intervals $S_i$ through backtracking, we first check \textit{SuccessCache}.
If an entry is found in the cache, we return the stored spatio-temporal profile as the optimal profile for the $(\phi_i, S_i)$ pair.
If no entry is found in \textit{SuccessCache}, we then check \textit{FailureCache}.
If an entry is found in \textit{FailureCache}, we declare `no solution' for the $(\phi_i, S_i)$ pair.
Otherwise, we call BCP to find spatio-temporal profiles.
If BCP returns a valid profile, then the $(\phi_i, S_i)$ pair along with the obtained profile is stored in \textit{SuccessCache}.
Otherwise, the $(\phi_i, S_i)$ pair is inserted into \textit{FailureCache}.

\section{MAMP on Grid Model}
In this section, we extend PSB to address MAMP on the grid model.
Given the three differences between the two models outlined in Sec.~\ref{sec:grid-def}, we make the following changes to PSB. 
Level 1 is applied to the grid model with no changes. However, the root PT node, in this case, contains no priority orderings because, as per Difference (D3), no two agents share the same start. 
While Level 2 can also be applied to the grid model with minor changes to handle Difference (D3), direct adoption results in poor scalability, as Difference (D1) causes SIPP to expand many duplicate SIPP nodes that reach the same collision point through different paths. To overcome this, we introduce a duplicate detection mechanism to prevent the expansion of duplicate nodes and a time window mechanism to expedite the planning process.
For Level 3, given that Lemma~\ref{lemma:exist} relies on the assumption of positive minimum speeds, which is no longer true due to Difference (D2), we incorporate a new method to initialize the upper bound. Additionally, due to Difference (D3), it is important to note that PSB is incomplete for the grid model. 

\subsubsection{Upper bound estimation for optimal arrival time}
In cases where $ub^m$ is infinite, we employ an exponential increment approach to establish the initial value for $T_{ub}$.
We begin by setting $T_{ub}$ to $T_{lb}$ and then evaluate the gradient of $\delta_{T_{ub}}^*$.
If the gradient is non-negative, then, as per Lemma~\ref{lemma:slack}, $T^* \in[T_{lb}, T_{ub}]$, in which case 
we find a valid upper bound and thus conclude the search process. Otherwise, we double $T_{ub}$ and evaluate its gradient again. 
However, if $T_{ub}$ reaches infinity (represented by the large value of $4000$ in our experiment), we declare `no solution' for the corresponding SIPP node.

\subsubsection{Duplicate detection} 
When extending Level 2 to the grid model, we may generate ``duplicate'' nodes that reach the same collision point through different paths.
Since the input of Level 3 contains the entire path and associated safe intervals from the root node to the current goal node, we cannot naively prune such ``duplicate'' nodes. 
Instead, we always add generated nodes to the open list even if nodes with identical collision points have been previously added. 
However, this method causes an exponential rise in the number of expanded nodes. 
Therefore, we develop a cleverer duplicate-detection method. Recall our reasoning on pseudo-identical paths for caching: if two goal nodes have pseudo-identical paths with exactly the same safe intervals, then their results from Level 3 should be identical. 
For example, in Fig.~\ref{fig:mapf_formulation} (b), if we consider the movement of agent 1, expanding the SIPP node at A generates two nodes at B and C. 
Expanding these nodes further results in two ``duplicate'' nodes at D.
While the two nodes correspond to two different paths from A to D, if their safe intervals are identical, they lead to the same LP models in Level 3. 
Thus, we can retain only one of the nodes without losing completeness.
Specifically, when we generate a node at collision point $c$, we prune it if we have previously generated a node that is also at $c$ and corresponds to a pseudo-identical path with exactly the same safe intervals.

\subsubsection{Windowed PSB} 
While the duplicate detection mechanism helps mitigate some duplicates, the SIPP search in Level 2 can still be time-consuming. We thus employ a rolling-horizon strategy\cite{fox1997dynamic,li2021lifelong}. 
In each iteration, we only focus on the collisions that occur within the current time window of size $t_W$.
Once PSB finds collision-free trajectories within the time window, we shift the window by a predetermined replan window $t_s < t_W$. (In our experiment, we set $t_s = 4, t_W = 6$.)
By recurrently replanning, we progressively advance the time window and compute the full trajectories for all agents.

    \begin{figure}[!t]
\centering
    \includegraphics[width=\linewidth]{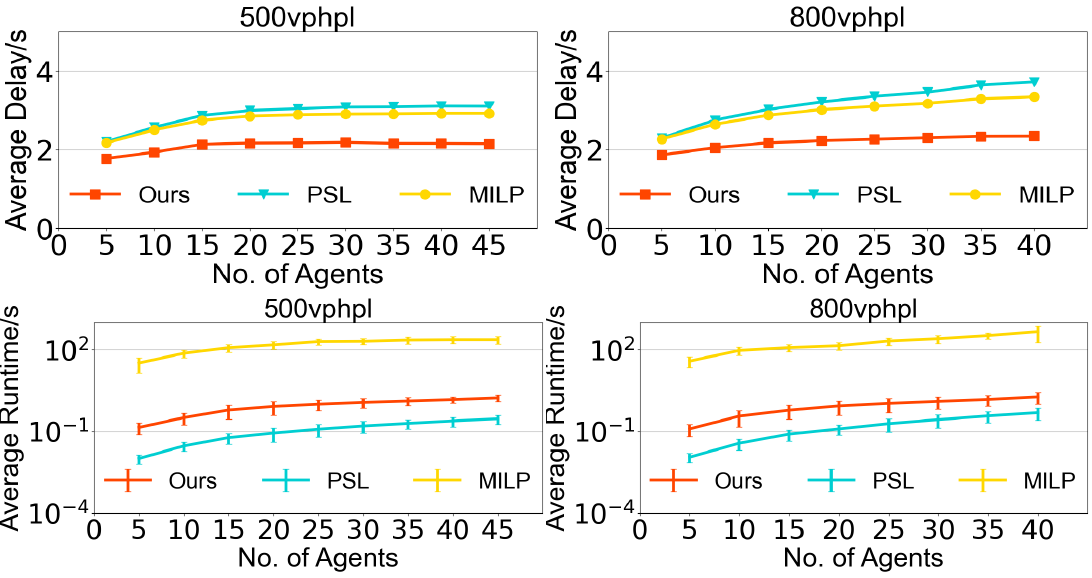}
    \caption{Delay and average runtime (in log scale) for the intersection model. 
    Error bars show the standard deviations.}
\label{fig:intersection_cost}
\end{figure}

\begin{figure*}[!t]
\centering
    \includegraphics[width=1.0\linewidth]{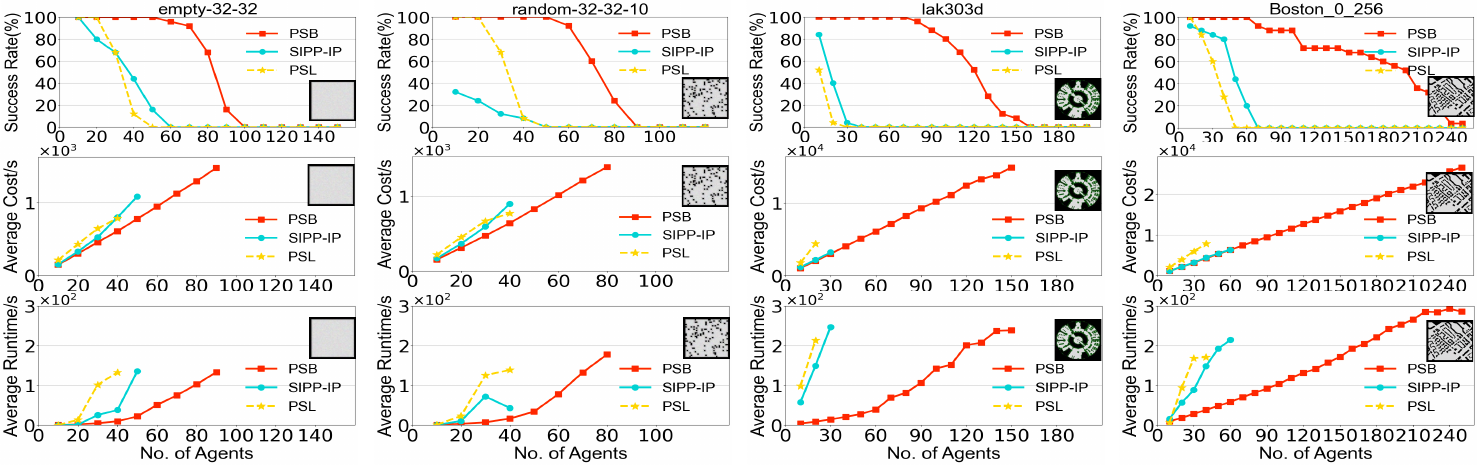}
    \caption{Success rate, solution cost, and runtime of PSB, PSL, and SIPP-IP across all maps for the grid model. 
    The solution cost and runtime are averaged only over scenarios where the planner successfully generates a solution. Note that PSL uses a relaxed agent model.}
\label{fig:mapf_sr}
\end{figure*}

\begin{figure}[!t]
\centering    \includegraphics[width=\linewidth]{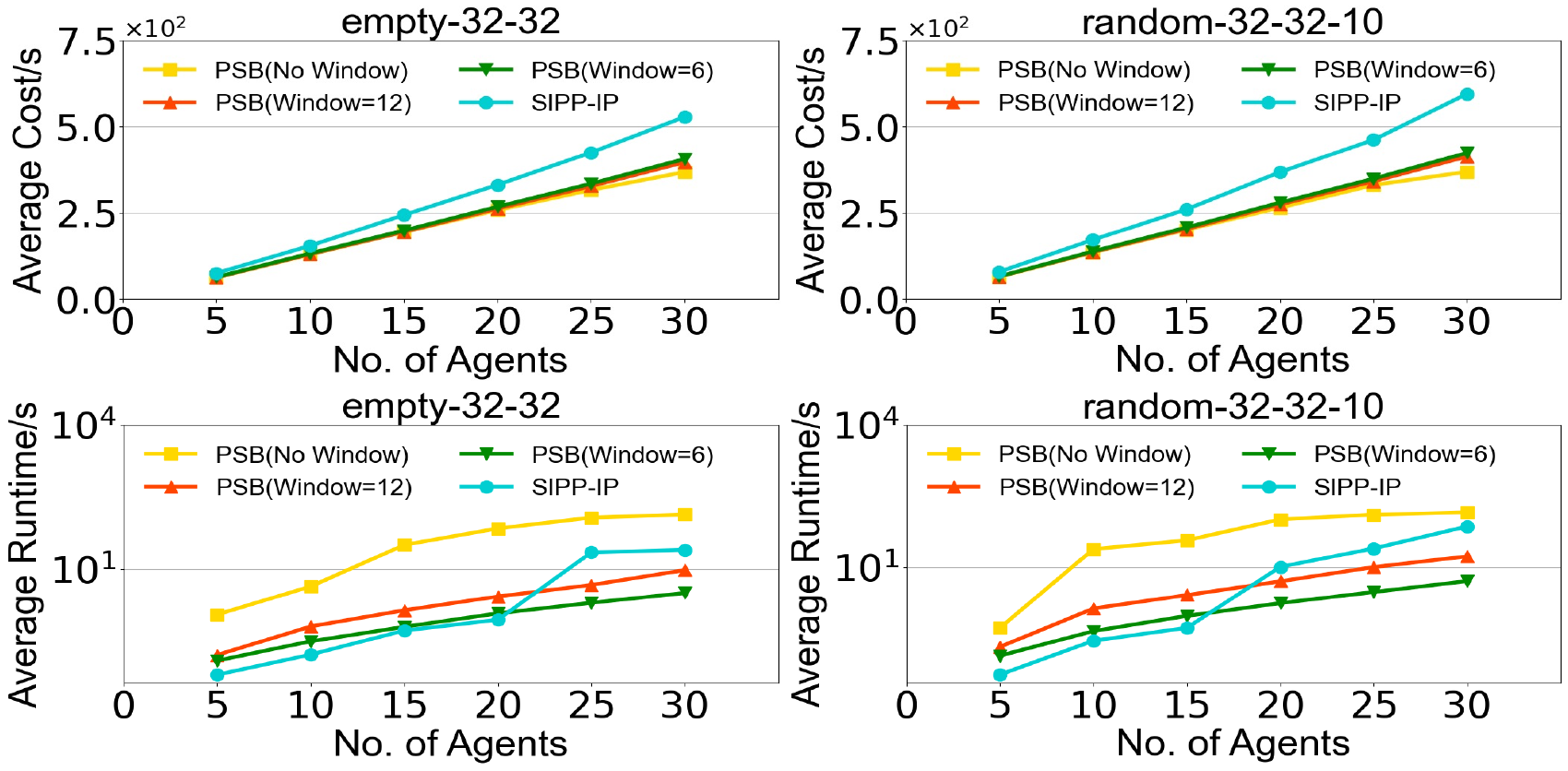}
    \caption{Runtime and cost of PSB with various window sizes.}
\label{fig:mapf_window}
\end{figure}

\begin{figure}[!t]
\centering    \includegraphics[width=\linewidth]{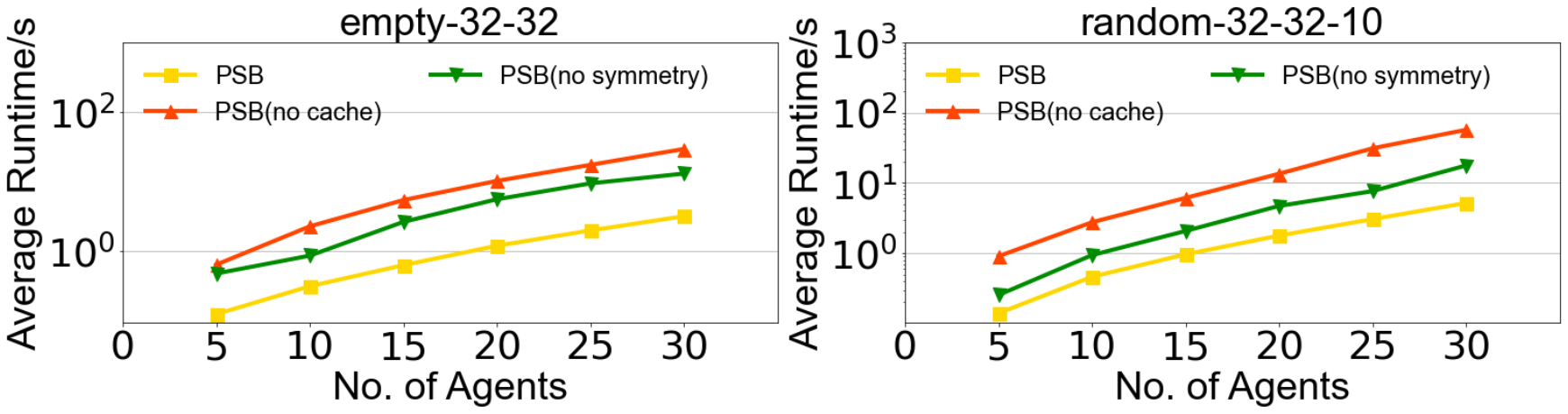}
    \caption{Runtime of PSB for ablation study.}
\label{fig:mapf_ablation}
\end{figure}

\section{Empirical Evaluation}
Both our PSB and baseline methods are implemented in C++ and utilize CPLEX to solve the programming models.\footnote{The source code for our method and baselines is publicly available at \repolink.} All experiments are conducted using a single core on an Ubuntu 20.04 machine equipped with an AMD 3990x processor and 188 GB of memory.

\subsection{MAMP on Intersection}
\subsubsection{Baseline Methods}
We include two baseline methods. The first baseline is PSL \cite{li2023intersection}, the previous algorithm that our PSB builds on. The second baseline is MILP \cite{levin2017conflict}, which solves the problem with mixed-integer linear programming. 
It is important to note that both methods assume that agents travel across the intersection at constant speeds.

\subsubsection{Simulation Setup}
We utilize a simulation setup identical to the one used by both baseline methods~\cite{levin2017conflict,li2023intersection}. 
Graph $G$ is depicted in Fig.~\ref{fig:mapf_formulation}. The lane width is $3.66\; m$, the right-turn radius is $1.83\; m$, and the left-turn radius is $9.14\; m$. The length of each agent is $L_i=5\; m$.
The entry lane for each agent is generated uniformly at random. An agent at an entry lane has an 80\% probability of going straight and a 20\% probability of turning left or right.
For simplicity, we only consider the speed and acceleration constraints.
We set a minimum speed of $\underline{U_i^1}=3\; m/s$ for all agents. 
The maximum speed for agents turning left is $\overline{U_i^1}=5\; m/s$, while that for other agents is $\overline{U_i^1}=15\; m/s$. 
As for acceleration, for all agents, we set $\underline{U_i^2}=-2\; m/{s^2}$ and $\overline{U_i^2}=5\; m/{s^2}$.
We randomly sample travel requests from two demands, namely 500 vphpl (i.e., 500 vehicles per hour per lane) and 800 vphpl, and report results averaged over 25 instances from each demand and each number of agents.


\subsubsection{Comparison}
While our objective is to minimize the sum of the arrival time for all agents, their arrival time depends on their earliest start time, which varies from instance to instance.
Thus, we instead report the \textit{average delay}, a popular metric used in transportation that measures the average difference between the arrival time and the \textit{earliest arrival time} (i.e., the earliest time an agent can reach its goal, defined by $e_i + d(c_i^{s}, c_{i}^{g})/\overline{U_i^1}$) among agents.
As shown in Fig.~\ref{fig:intersection_cost}, while PSB runs slower than PSL, PSB shows an improvement of up to 41.15\% for 500 vphpl demands and 49.79\% for 800 vphpl demands compared to PSL in terms of average delay.
Although MILP is optimal under constant speed assumption\cite{levin2017conflict}, it still produces worse solutions than PSL.
In terms of runtime, MILP is significantly slower than both PSL and PSB.
At the same time, the difference between the average delays of PSL and MILP is much smaller than that between PSB and PSL.
These improvements arise from the fact that the baseline methods assume the constant speed during intersection traversal, while our planner considers the full kinodynamic capabilities of agents.
We also notice that, in the 500 vphpl scenario, the average-delay curves of all three methods level off as the number of agents increases. 
In contrast, in the 800 vphpl scenario, only the delay curve of PSB levels off, suggesting that PSB can find a stable solution for higher demand scenarios.

\subsection{MAMP on Grid Model}
\subsubsection{Baseline Methods}
In this experiment, we compare PSB with PSL and a straightforward extension of SIPP-IP~\cite{ali2023safe}.
SIPP-IP is a state-of-the-art single-agent path planner designed to accommodate kinodynamic constraints, making it a suitable representation of motion-primitive-based methods.
The same kinodynamic motion primitives as \cite{ali2023safe} are used during the evaluation.
To adapt SIPP-IP for multi-agent scenarios, we replaced Level 2 and Level 3 in PSB with the SIPP-IP. 

\subsubsection{Simulation Setup}
We evaluate PSB, PSL, and SIPP-IP on four four-neighbor grid maps from the MAPF benchmark \cite{Stern2019benchmark}, namely \texttt{empty} (size: 32$\times$32), \texttt{random} (size: 32$\times$32), \texttt{lak303d} (size: 194$\times$194), and \texttt{Boston} (size: 256$\times$256).
For each map, we conducted experiments with a progressive increment in the number of agents, using an average of 25 random instances from the benchmark set. 
The agents are modeled as disks with a diameter of $0.99\; cell$.
All agents have identical kinodynamic constraints, which, similar to the intersection model, encompass only speed and acceleration constraints.
The speed is bounded by the range of $[0, 2]\; cell/s$, while the acceleration is confined to $[-0.5, 0.5]\; cell/s^2$. 
To be noticed, since PSL assumes agents move at a constant speed, we have relaxed the acceleration constraints on starts and goals for PSL, allowing agents to adjust to the desired speed immediately.

\subsubsection{Comparison}
We evaluate solution quality using the sum of the arrival time of all agents.
As shown in Fig.~\ref{fig:mapf_sr}, PSB shows better solution quality than both PSL and SIPP-IP while achieving comparable or better runtime performance.
PSB also outperforms them across all four maps in terms of \textit{success rate}, which represents the ratio of instances solved within $300 s$ over all instances.
This can be attributed to two factors. 
Firstly, the longer runtime of PSL and SIPP-IP in scenarios with more agents makes it easier to hit the preset cutoff time.
Secondly, the solution from baseline methods is limited in action choices, leading to failures in solving certain cases.
For instance, when $c_i^g$ is adjacent to the $c_i^s$, SIPP-IP encounters difficulties in finding an appropriate solution (The accelerating primitive defined in \cite{ali2023safe} takes 4 cells).

\subsubsection{Ablation study}
We first evaluate the influence of window size $t_W$ by comparing our method under different window sizes: $t_W=6$, $t_W=12$, and $t_W=\infty$. 
As shown in Fig.~\ref{fig:mapf_window}, we observe marginal improvement in solution quality as we increase $t_W$ (less than 5\% from $t_W=6$ to $t_W=\infty$), while causing a significant rise in runtime.
Furthermore, as shown in Fig.~\ref{fig:mapf_ablation}, PSB exhibits a runtime improvement of up to 88.58\% compared to PSB without the cache mechanism and 76.71\% improvement over PSB without the duplicate detection mechanism.
Importantly, these two mechanisms do not affect the solution quality of the algorithms.
    \section{Conclusion}
This paper introduces PSB, a three-level planner designed to tackle MAMP with kinodynamic constraints. 
PSB produces smooth solutions by effectively utilizing the full kinodynamic capacity of agents, addressing a limitation often faced by existing methods. 
We apply PSB to two domains: traffic intersection coordination for autonomous vehicles and obstacle-rich grid map navigation for mobile robots.
In both domains, PSB outperforms the baseline methods with up to 49.79\% improvement in terms of solution quality, while achieving better success rates and comparable runtime. 

    \bibliographystyle{IEEEtran}
    \bibliography{reference}
\endgroup
\end{document}